\documentclass[journal]{IEEEtran}
\usepackage{blindtext}
\usepackage[pdftex]{graphicx}
\graphicspath{{images/}}
\DeclareGraphicsExtensions{.pdf}
\usepackage[export]{adjustbox}
\usepackage{algpseudocode}
\usepackage{algorithm}
\algnewcommand\algorithmicforeach{\textbf{for each}}
\algdef{S}[FOR]{ForEach}[1]{\algorithmicforeach\ #1\ \algorithmicdo}

\let\oldReturn\Return
\renewcommand{\Return}{\State\oldReturn}
\usepackage{subcaption}
\usepackage{color,soul}
\usepackage{etoolbox}
\usepackage[cmex10]{amsmath}
\begin{document}

\title{EVM-CNN: Real-Time Contactless Heart Rate Estimation from Facial Video}

\author{Ying~Qiu, Yang Liu, Juan Arteaga-Falconi, Haiwei Dong,~\IEEEmembership{Senior Member,~IEEE,} and Abdulmotaleb El Saddik,~\IEEEmembership{Fellow,~IEEE}
\thanks{Y. Qiu, L. Yang, J. Arteaga-Falconi, H. Dong, and A. El Saddik are with Multimedia Computing Research Laboratory (MCRLab), School of Electrical Engineering and Computer Science, University of Ottawa, Ottawa, ON, K1N 6N5 Canada e-mail: \{yqiu059; yliu344; jarte060; hdong; elsaddik\}@uottawa.ca}
}
\maketitle

\begin{abstract}
With the increase in health consciousness, non-invasive body monitoring has aroused interest among researchers. As one of the most important pieces of physiological information, researchers have remotely estimated the heart rate from facial videos in recent years. Although progress has been made over the past few years, there are still some limitations, like the processing time increasing with accuracy and the lack of comprehensive and challenging datasets for use and comparison. Recently, it was shown that heart rate information can be extracted from facial videos by spatial decomposition and temporal filtering. Inspired by this, a new framework is introduced in this paper for remotely estimating the heart rate under realistic conditions by combining spatial and temporal filtering and a convolutional neural network. Our proposed approach shows better performance compared with the benchmark on the MMSE-HR dataset in terms of both the average heart rate estimation and short-time heart rate estimation. High consistency in short-time heart rate estimation is observed between our method and the ground truth.
\end{abstract}

\IEEEpeerreviewmaketitle

\section{Introduction}
The human heart rate (HR) reflects both physiological and psychological conditions. HR monitoring has benefits for human beings in many aspects, such as health care for elders, vital sign monitoring of newborns and lie detection in criminals. Nowadays, with the growing health consciousness, mobile HR monitoring systems are becoming increasingly popular with users. However, all such devices must be in contact with the skin, which is inconvenient and may cause discomfort. Hence, researchers have been devoted to creating a low-cost and noninvasive way to measure HR in recent years. The main concepts are derived from the principles of photoplethysmography (PPG), which can sense the cardiovascular blood volume pulse through variations in transmitted or reflected light \cite{allen2007photoplethysmography}. Furthermore, Verkruysse \textit{et al}. showed that a PPG signal can be measured using a standard digital camera with ambient light as an illumination source \cite{verkruysse2008remote}. Since then, researchers have made sustainable progress in digital-camera-based remote heart rate estimation.

However, various challenges remain. Because the face is the least occluded skin region of the human body, extracting a PPG signal from the facial region has become the main approach \cite{hassan2016optimal}. The motion artifacts of the subject affect the performance of measurements, which are divided into two types: One is rigid motion, which includes head tilt and posture changes. The other is non-rigid motion, which includes facial expressions such as eye blinking and smiling. Illumination variations also add noise to the PPG signal, such as the flash of indoor lights, the variation of reflected light from a computer screen and the internal noise of the digital camera \cite{li2014remote}. In addition, other challenges affect this technology, such as signal strength and a lack of appropriate datasets.

The approaches proposed by researchers in recent years can be divided into two groups. The methods in the first group select the whole face as the region of interest. Then, most of the methods use a filter to reduce the noise caused by motion artifacts and illumination variations. The methods in the other group focus on choosing a region of interest (ROI) of the face and estimating the HR based on this reliable region. To some extent, both methods have aspects of superiority, but downsides do exist.

Whole-face-region-based approaches usually extract a PPG signal by spatially averaging the face region at the outset and then mainly focusing on reducing the impact of interference by the noise. Such approaches, e.g., \cite{li2014remote} and \cite{poh2011advancements}, perform better under non-rigid motion conditions than they do under rigid motion conditions because computing the mean of the whole face nullifies the effect of facial muscle movements. However, the whole face region always includes noise caused by multiple environmental factors. Although various filters are applied to eliminate the noise, the processing time increases with the number of filters being used. Thus, it is difficult to estimate HR instantaneously by using such an approach.

Partial face region selection can help decrease the time spent filtering the signal. Reliable region selection is a challenge in such approaches since it contains too much indeterminacy, particularly when the subject moves with facial expressions during the experiment. Progress has been made, such as in \cite{tulyakov2016self} and \cite{lam2015robust}, with better performance than the whole-face-region-based approach.

Regardless of which of the above methods are considered, there is a common point among them: they all use the power spectral density of the underlying signal to estimate the HR in the last step of the procedure, which requires a clear PPG signal to obtain an accurate result. Many processing steps must be used to obtain such a clear signal, which increases the computing time. To acquire better performance in less time, additional knowledge must be considered. Inspired by deep learning, HR estimation has been one of the hottest topics in recent years and has attracted researchers from a variety of fields, who have used it to cope with various tasks, such as \cite{zhang2016joint}, \cite{lu2015rating}, and \cite{zhang2016deep}. HR estimation is considered a regression task in this paper. Because subtle changes in skin color related to cardiac rhythm can be extracted from a digital camera, the average HR within a short time interval can be defined as a label in the regression task, and the extracted color changes of the corresponding video sequence can be fed into the neural network. In the research of Wu \textit{et al.}, the blood flow through the face due to the cardiac cycle is amplified as the skin color changes, and the vertical scan from the amplified video sequence reveals a plausible human heart rate \cite{wu2012eulerian}. Inspired by this, a new paradigm is proposed for estimating the heart rate by using Eulerian Video Magnification (EVM) to extract face color changes and using deep learning to estimate the heart rate. 

The dataset is one of the most important factors that can affect the performance of deep learning. In addition, the dataset can be used to make a fair comparison of different approaches. Although the MAHNOB-HCI dataset \cite{soleymani2012multimodal} is widely used in HR estimation research, there are several limitations that must be considered. The experiment conditions are strictly controlled, and the subjects hardly move and do not show obvious facial expressions. To achieve high generalization ability under realistic conditions, we perform our method on the MMSE-HR dataset \cite{zhang2016multimodal}, which is more challenging because of its larger multimodal spontaneous emotion corpus.

In our work, EVM and deep learning are integrated to achieve instantaneous HR estimation. Specifically, EVM is used to extract the feature image that corresponds to the heart rate information within a time interval. A convolutional neural network (CNN) is then applied to estimate HR from the feature image, which is formulated as a regression problem. In summary, the contributions of this paper are as follows:

\begin{enumerate}
    \item  A new paradigm is proposed for HR estimation through a digital camera under realistic conditions. As part of EVM, spatial decomposition and temporal filtering are applied to extract face color changes as the feature image. A wider bandpass filter is used to extract the signal within the general range of the human HR.
    \item  HR estimation is achieved by using a CNN, where the input is the feature image within a specified time interval and the output is the average HR during that time interval. Our approach addresses the problems of high computational complexity and time cost.
    \item  To demonstrate the performance of our approach, a comparison between the proposed approach and a benchmark \cite{tulyakov2016self} on the same dataset for average HR estimation and short-time HR estimation was conducted. Our approach achieves higher accuracy than the other methods.

\end{enumerate}

The rest of the paper is organized as follows: In Section II, a review of the state-of-the-art methods for remote HR estimation is presented. The details of the proposed method are described in Section III. Then, in Section IV, the dataset, evaluation metrics and experimental settings are discussed. In Section V, results and analysis are presented. Finally, the conclusions of our work and directions for future work are discussed in Section VI.

\begin{figure*}[t]
    \centering
    \includegraphics[width=\textwidth]{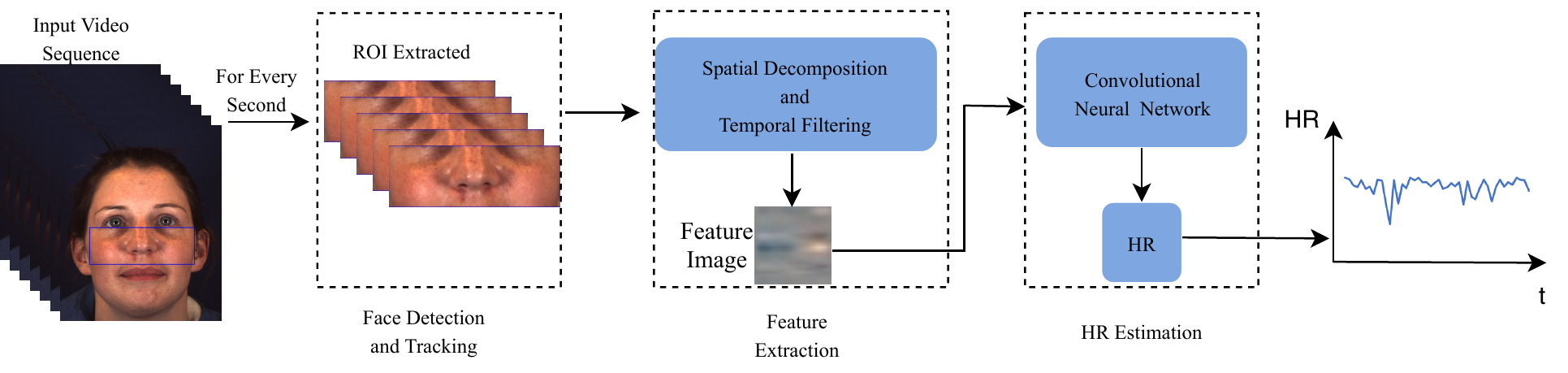}
    \caption{Diagram for HR estimation. In the face detection and tracking module, regions of interest are automatically extracted and input into the next module. Spatial decomposition and temporal filtering are applied to obtain the feature image in the feature extraction module. In the HR estimation module, a CNN is used to estimate the HR from the feature image.}
    \label{diagram}
\end{figure*}

\section{Related Work}
Traditional measurement methods for cardiac activity are contact-based methods and include extracting the electrocardiography (ECG) signal by using an ECG sensor and the PPG signal by using a pulse oximeter. Although contact-based methods can measure cardiac activity comprehensively with high accuracy, sensors that are attached to human bodies always cause discomfort and inconvenience. Researchers have made progress in creating a noninvasive measurement method over the past few years.

In 2009, Verkruysse \textit{et al.} showed that a PPG signal can be measured remotely by using ambient light and a simple consumer-level digital camera in movie mode \cite{verkruysse2008remote}. It is also shown that the green channel features the strongest PPG signal since the hemoglobin light absorption reaches its peak in the green channel; the red and blue channels also contain PPG information. In 2010, Poh \textit{et al.} introduced a non-contact cardiac pulse measurement method with color images that uses blind source separation \cite{poh2010non}. They computed the spatial average value of the face region for each channel, where three temporal traces were obtained from an RGB video sequence. Independent component analysis was applied to separate the observed raw signals and identify the underlying PPG signal. Finally, they applied Fast Fourier Transform (FFT) on the selected source signal to find the highest power spectrum within the general human HR frequency band and deal with it as the average HR frequency of the video recording. In 2011, Poh \textit{et al.} improved their work by adding several temporal filters to clean up the PPG signal \cite{poh2011advancements}. 

In 2012, Wu \textit{et al.} introduced Eulerian Video Magnification, which can amplify the blood flow on the human face in a video recording \cite{wu2012eulerian}. Each pixel of the face region was processed by spatial decomposition and temporal filtering. Then, the resultant frames were multiplied by a magnification factor and added to the original frames. Therefore, the subtle color changes due to the cardiac cycle are visible to the naked eye. In 2013, Balakrishnan \textit{et al.} showed that HR can be extracted from videos by measuring subtle head motions caused by a Newtonian reaction to the influx of blood at each beat \cite{balakrishnan2013detecting}. They used principal component analysis to decompose the trajectories and choose the one whose temporal power spectrum best matched the pulse, from which the average HR was calculated.

After several methods were successfully introduced for estimating HR remotely, researchers began to focus on this topic under challenging conditions. In 2013, De Haan \textit{et al.} presented an analysis of the limitations of blind source separation on motion problems and proposed a chrominance-based approach that achieved better performance \cite{de2013robust}. Li \textit{et al.}, in 2014, proposed a framework for estimating HR under realistic conditions \cite{li2014remote}. They used face tracking to solve the problem of rigid movements and employed an adaptive filter to rectify the impact of illumination variations. They also compared their performance with those of existing methods on the MAHNOB-HCI dataset \cite{soleymani2012multimodal}; their approach showed higher accuracy than did the other methods. In 2015, Lam \textit{et al.} improved blind source separation by randomly choosing a pair of patches from the face, obtained multiple extracted PPG signals, and, finally, used a majority voting scheme to robustly recover the HR \cite{lam2015robust}. Different from existing methods based on RGB videos, Yang \textit{et al.} proposed a non-intrusive heart rate estimation system via 3D motion tracking in depth video \cite{yang2017estimating}. In 2018, Prakash \cite{prakash2018bounded} used a bounded kalman filter for motion estimation and feature tracking in remote photo PPG methods. To tackle the blurring and noise due to random head movements, a blur identification and denoising are employed for each frame.

Although the accuracy was improved and the system became more robust, the processing time was still a drawback, which was too long for estimating HR instantaneously. Tulyakov \textit{et al.} used a self-adaptive matrix completion approach to automatically discard the face regions corresponding to noisy features and proposed using only reliable regions to estimate HR \cite{tulyakov2016self}. They also introduced a more challenging dataset with target movements and spontaneous facial expressions. Moreover, instantaneous HR can be estimated by using their method. 

In 2017, Alghoul \textit{et al.} compared two methods \cite{alghoul2017heart}: one is based on independent component analysis (ICA), and the other is based on EVM. They concluded that the ICA-based method showed better results in general but that the EVM-based method might be more appropriate when motion is involved. Inspired by this, a new framework is introduced that combines EVM and deep learning to estimate HR instantaneously. Instead of applying the magnification process of EVM, spatial decomposition and temporal filtering are mainly used to extract the feature image as described in the next section. 

\section{Method}
\label{ME}
In this section, an overview of the proposed method is presented through three procedures, as illustrated in Fig. \ref{diagram}. The face in the input video sequence is detected and tracked to extract the ROI. Then, the extracted ROI of each frame is processed by using spatial decomposition and temporal filtering to obtain the feature image. Finally, the feature image is input into a convolutional neural network to estimate the HR.

\subsection{Face Detection and Tracking}
\label{face_detect}
Face detection and tracking are used to extract the face regions from the images and keep the size of the face within a certain time range. The proposed approach is designed for practical conditions and must automatically select the ROI and feed it into the feature extraction module. Under practical conditions, several factors may impact face region extraction, such as the subject's movements, head rotations and facial expressions. To obtain a steady face region without non-skin pixels, precise facial landmarks are used to define a face region.

\begin{figure}[h!]
    \centering
    \includegraphics[width=0.4\textwidth, center]{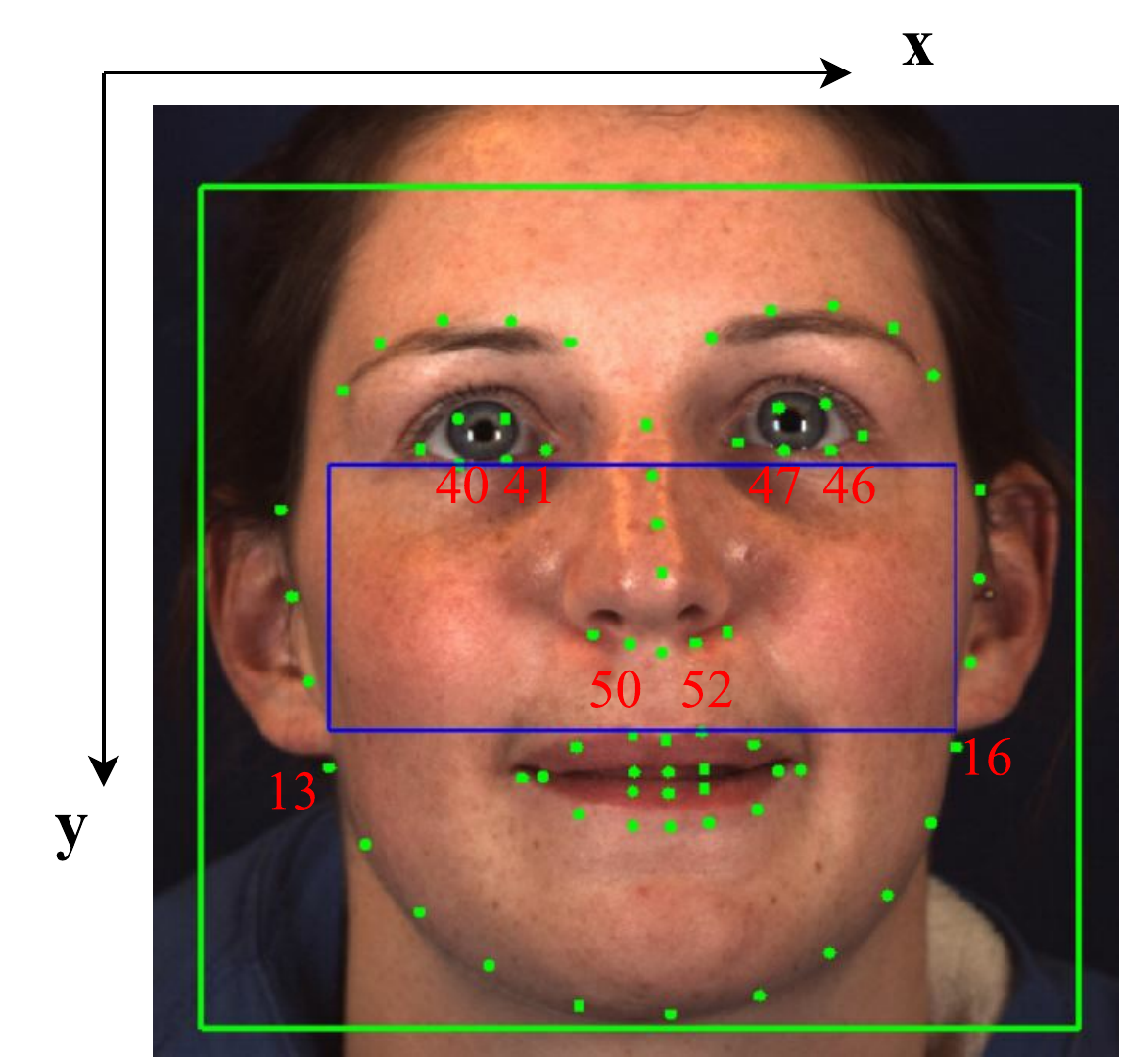}
    \caption{Face detection and facial landmarks. The green rectangle denotes the bounding box of the face detection result. Facial landmarks are denoted by 68 green points. The blue rectangle indicates that the ROI is defined by 8 points, which are denoted by the nearby red numbers.  }
    \label{face_landmark}
\end{figure}

A regression local-binary-features-based approach is applied to detect the bounding box and 68 facial landmarks inside the bounding box \cite{ren2014face}. Local binary features for each landmark are first generated by learning a local feature mapping function. All the features are concatenated and a linear projection is learned by linear regression. This process is repeated stage-by-stage in a cascaded fashion. To define the ROI, 8 points are used, as shown in Fig. \ref{face_landmark}. The green rectangle is the face detection result, and the 68 facial landmarks are represented by 68 green points. The blue rectangle is the ROI defined by the coordinates of the 8 points that are denoted by red numbers nearby. The directions of the $x$- and $y$-axes are also shown in Fig. \ref{face_landmark}. The top-left vertex coordinate and the size of the blue rectangle are defined in Eq. 1:

\begin{equation}
    \begin{cases}
    X_{LT}=X_{P_{13}} \\
    Y_{LT}=\max(Y_{P_{40}}, Y_{P_{41}}, Y_{P_{46}}, Y_{P_{47}}) \\
    W_{rect}=X_{P_{16}}-X_{P_{13}} \\
    H_{rect}=\min(Y_{P_{50}}, Y_{P_{52}})-Y_{LT} \\
    \end{cases}
\end{equation}
where $X_{LT}$ and $Y_{LT}$, respectively, denote the $x$ and $y$ coordinates of the top-left vertex; $X_{P_{13}}$ denotes the $x$ coordinate of Point 13; $Y_{P_{40}}$, $Y_{P_{40}}$, $Y_{P_{41}}$, $Y_{P_{46}}$, and $Y_{P_{47}}$ denote the $y$ coordinates of Point 40, Point 41, Point 46, and Point 47, respectively; $W_{rect}$ denotes the width of the blue rectangle; $X_{P_{16}}$ and $X_{P_{13}}$ denote the $x$ coordinates of Points 16 and 13, respectively; $H_{rect}$ denotes the height of the blue rectangle; and $Y_{P_{50}}$ and $Y_{P_{52}}$ denote the $y$ coordinates of Points 50 and 52, respectively. Since the blue rectangle is defined in this way, the ROI always excludes the eye part and the mouth part, regardless of how the subjects rotate their heads. Furthermore, the impact of eye blinking and mouth movements is reduced. Hence, a cheek region without non-facial pixels is defined.

As the input to the feature extraction module, the fixed-size ROI rectangle must be within a certain time interval because the feature extraction module must obtain the color changes of each fixed pixel over time. To achieve this, the Scalable Kernel Correlation Filter Tracking method \cite{montero2015scalable} is applied to track the ROI rectangle and keep the ROI's size constant within a certain time interval, which can reduce the impact of rigid head motion. Then, the tracked cheek region is input into the feature extraction module to obtain the skin color changes during the cardiac cycle.

\subsection{Feature Extraction}
\label{feature_extract}
\begin{figure}[h!]
    \centering
    \includegraphics[width=0.5\textwidth, center]{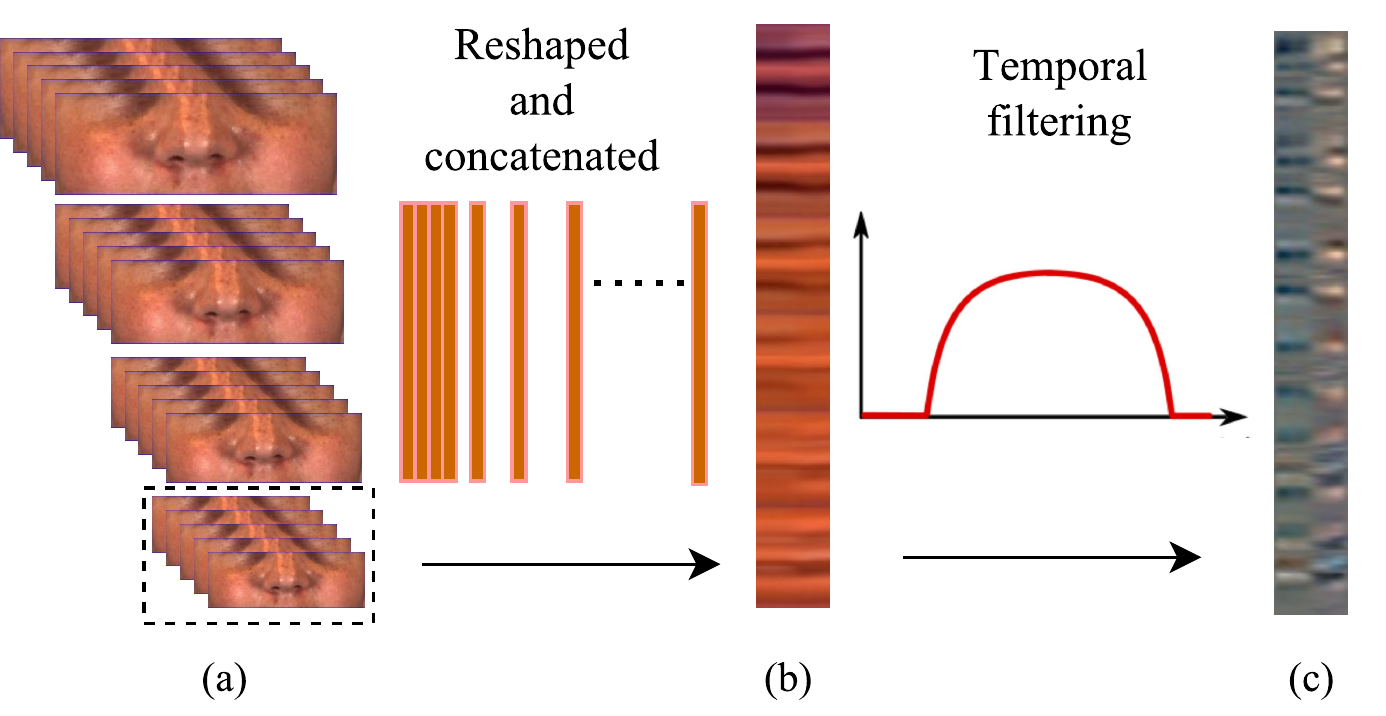}
    \caption{Feature extraction. (a) shows the module that decomposes the input sequence of ROIs into multiple spatial frequency bands, (b) demonstrates the result after reshaping and concatenating the lowest band of (a), which is marked by a black dashed rectangle, and (c) is the feature image obtained by temporal filtering of the concatenated image with a frequency band of interest.}
    \label{evm}
\end{figure}

Wu \textit{et al.} consider the time series of color values at any spatial pixel and amplify the variation in a given frequency band of interest \cite{wu2012eulerian}. Inspired by this, the blood flow information extracted from a video sequence is applied in our approach. By using localized spatial pooling and temporal filtering to extract the signal that corresponds to the pulse, they multiply the extracted signal with a magnification factor to amplify the facial color changes over time and make them visible to the naked eye. Inspired by this, spatial decomposition and temporal filtering are applied in our approach to extract the feature image that contains the signal related to the HR information. An overview of spatial decomposition and temporal filtering is given in Fig. \ref{evm}. As shown in Fig. \ref{evm}(a), the input sequence is first decomposed into multiple spatial frequency bands. Then, the lowest-band sequence is reshaped and concatenated to obtain a new image, as shown in Fig. \ref{evm}(b). Finally, a bandpass filter is used to obtain the feature image that contains the signal related to blood flow, as shown in Fig. \ref{evm}(c).

Spatial decomposition for color magnification is implemented by using Gaussian pyramid decomposition. Each frame of the input RGB video is downsampled to a specified level. The last layer of each downsampled frame, which is shown as the part in the black dashed rectangle in Fig. \ref{evm}(a), is reshaped into one column. All columns obtained in the previous step are concatenated to obtain a new image. To detect instantaneous HR, a one-second interval video sequence is input into the feature extraction module for every round rather than the whole video. Hence, the number of columns of the concatenated image in Fig. \ref{evm}(b) should be equal to the number of frames per second.

Temporal filtering is the use of an ideal bandpass filter to obtain the signal with the frequency of interest. Different from \cite{wu2012eulerian}, a wider bandpass filter is used here to cope with typical conditions. A human's HR is 45-240 bpm. Thus, the corresponding frequency band is computed as $f_{HR}=HR/60$, which is 0.75-4.0 Hz. As explained before, each column of the concatenated image corresponds to a downsampled image of the ROI and each row of the concatenated image corresponds to the variations of a fixed position (pixel) in the ROI over one second. Therefore, each channel of the concatenated image is transferred to the frequency domain by using FFT by rows. Then, a mask of the same size is used to retain the component within the frequency band of interest and set the others to zero. Finally, Inverse Fast Fourier Transform (IFFT) is performed by rows to transform the signal back into the time domain, and three channel images are merged to obtain the feature image, as shown in Fig. \ref{evm}(c). The spatial decomposition and temporal filtering algorithm is named Feature Extraction and summarized in Algorithm 1.

\begin{algorithm}[h!] 
\caption{Feature Extraction}
\begin{algorithmic}[1]
\Require The original RGB video frames $I$, pyramid level $Pl$, frame rate $Fps$, set of one-column intermediate images $C$, low-frequency cut-off $Fl$ and high-frequency cut-off $Fh$ of the ideal bandpass filter
\Ensure A set of feature images $S$.
\Repeat
\ForEach {frame $I$}
\State Gaussian pyramid with level $Pl$
\State Reshape the last level to one column $C_{0}$
\EndFor
\State $C \gets C_{0}$
\Until {$size(C)=Fps$}
\Repeat
\ForEach {set $C$}
\State Concatenate all the columns into a new image $M$
\ForEach{channel of $M$}
\State Do FFT to obtain $Mf$
\State Create a mask by using $Fl$ and $Fh$
\State Multiply the mask and $Mf$ to get $N$
\State Do IFFT of $N$ to obtain $Ni$
\EndFor
\State merge 3 channels to obtain $K$
\EndFor
\State $S \gets K$
\Until {the whole video ends}
\Return $S$
\end{algorithmic}
\end{algorithm}

The ROI frame is first downsampled into different bands, the last level is reshaped to one column which is stored into a vector until the number of the elements is equal to the frame rate. All the columns are then concatenated orderly to get a new image, and the new image is split into three channels. Each channel is transferred to frequency domain and multiplied with a mask to zero out the FFT components outside the frequency band of interest. Finally, 3 channel are transferred back to time domain and merged to obtain the feature image. The feature extraction module is a loop for each second video frames and it stops until the video ends.

\subsection{HR Estimation by CNN}
\label{cnn}
\begin{figure}[h!]
    \centering
    \includegraphics[width=0.48\textwidth,left]{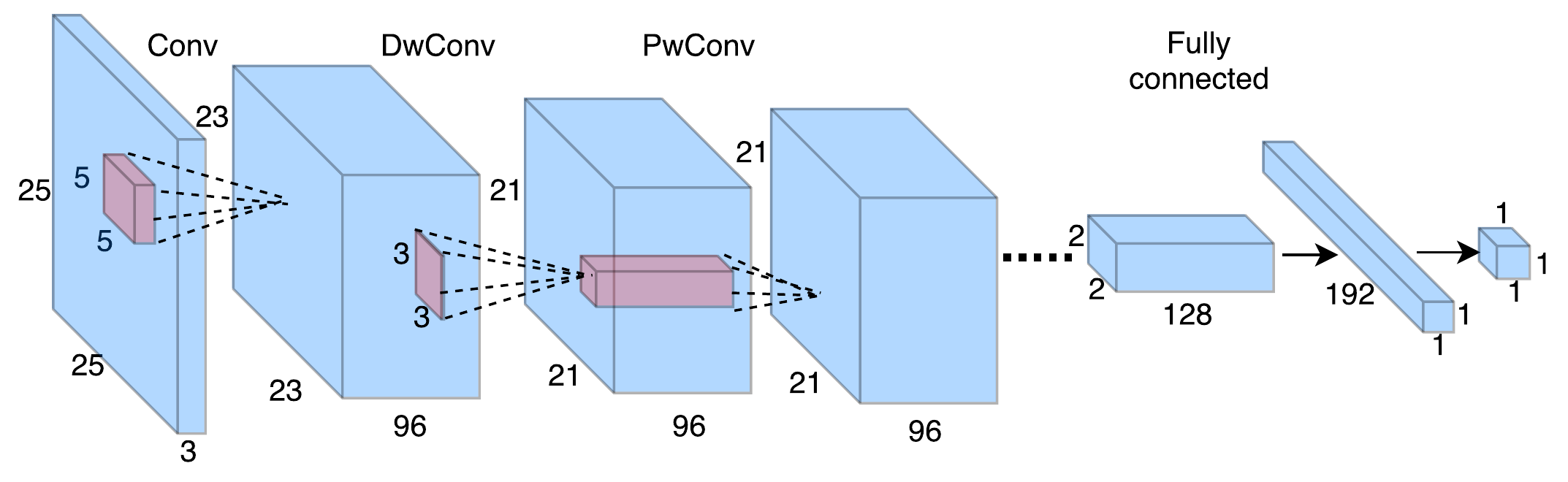}
    \caption{Structure of the convolutional neural network, where for each layer, the light blue cube denotes the input, the light pink cube denotes the reception field, the size of each cube is labeled beside it, the relative number of channels are represented by the thickness, and the arrow denotes the data flow. }
    \label{cnn4}
\end{figure}

HR is estimated from the temporal image extracted in the previous procedure. This is achieved by using a regression convolutional neural network. A feature image is input into the network, and a corresponding HR is the output. The pipeline of the neural network is designed as shown in Fig. \ref{cnn4}. Considering the frame rate of the video sequence, which is related to the number of columns of the feature image, the input image is defined to be of size 25$\times$25$\times$3. Inspired by MobileNet \cite{howard2017mobilenets}, depthwise separable convolutions are used as the main structure in this part. Depthwise separable convolution is a form of factorized convolution that is used to factorize a standard convolution into a depthwise convolution and a 1$\times$1 pointwise convolution, which can effectively reduce the computational burden and model size.

\begin{table}[h!]
\centering
\caption{CNN Body Architecture}
\resizebox{0.48\textwidth}{!}{
\begin{tabular}{c|c|c}
\hline
Input Size & Type/Stride & Filter Shape \\
\hline
$25\times25\times3$ & Conv/s1 & $5\times5\times3\times96$ \\
\hline
$23\times23\times96$ & DwConv/s1 & $3\times3\times96$ dw \\
\hline
$21\times21\times96$ & PwConv/s1 & $1\times1\times96\times96$ \\
\hline
$21\times21\times96$ & DwConv/s2 & $3\times3\times96$ dw \\
\hline
$11\times11\times96$ & PwConv/s1 & $1\times1\times96\times96$ \\
\hline
$11\times11\times96$ & DwConv/s2 & $3\times3\times96$ dw \\
\hline
$6\times6\times96$ & PwConv/s1 & $1\times1\times96\times128$ \\
\hline
$6\times6\times128$ & DwConv/s2 & $3\times3\times128$ dw \\
\hline
$3\times3\times128$ & PwConv/s1 & $1\times1\times128\times128$ \\
\hline
$3\times3\times128$ & DwConv/s2 & $3\times3\times128$ dw \\
\hline
$2\times2\times128$ & PwConv/s1 & $1\times1\times128\times128$ \\
\hline
$2\times2\times128$ & AvePool/s1 & Pool $2\times2$ \\
\hline
$1\times1\times192$ & FC/s1 & $128\times192$ \\
\hline
$1\times1\times192$ & Dropout/s1 & ratio $0.6$ \\
\hline
$1\times1\times192$ & FC/s1 & $192\times1$ \\
\hline
$1\times1\times1$ & Eu/s1 & Regression \\
\hline
\end{tabular}
}
\label{table:cnn}
\end{table}

As shown in Fig. \ref{cnn4}, the first layer is a full convolution layer, which is denoted ``Conv"; for the next layer, a depthwise convolution layer is used to filter each input channel, and a pointwise layer is used to combine the outputs of the depthwise convolution layer, which are denoted ``DwConv" and ``PwConv", respectively. The first full convolution layer is obtained by applying a kernel of size 5$\times$5$\times$3$\times$96, where 5$\times$5 is the height and width of the filter; 3 represents the depth, which corresponds to the number of input channel (RGB); and 96 is the number of output feature maps. In the following depthwise convolution layer, a kernel of size 3$\times$3$\times$96 is used to filter the input feature map, where 3$\times$3 denotes the height and width and 96 is the number of input channels. A hidden parameter of this filter is the depth, which is always equal to 1 due to the property of depthwise convolution. Similarly, the size of next pointwise filter is $1\times1\times96\times96$, where 1$\times$1 is the height and width, the first 96 is the depth of the filter and the second 96 is the number of outputs. In the following parts, which are represented by an ellipsis, a similar depthwise separable convolution structure is used. The details of the CNN Body architecture are illustrated in Table \ref{table:cnn}. Each convolution layer is followed by a batch normalization and ReLU nonlinearity. Downsampling is achieved by changing the value of the stride. An average pooling layer downsamples the final feature map to 1 and computes the average value of each channel. It simply performs downsampling along the spatial dimensionality of the given input. By having less spatial information, not only will the computation performance be improved, the probability of overfitting will also be reduced. A fully connected layer is applied to directly connect every neuron in one layer to every neuron in another layer. A dropout layer is then used to avoid the overfitting problem. The dropout radio is set to 0.6, which can stop 60\% of the feature detectors from working during the process of training and improve the generalization of the network capacity.

Since the HR of each feature image is the only label, the last fully connected layer has one neuron. All the labels are normalized to the range 0-1 before training, where 0 corresponds to 45 bpm and 1 corresponds to 240 bpm. The Euclidean distance is chosen as the loss function for measuring the difference between the predicted value and the ground truth. The equation is defined as Eq. 2:

\begin{equation}
L_{Eu}=\frac{1}{2N} \sum_{i=1}^{N}\|q_i-p_i\|    
\end{equation}
where $L_{Eu}$ denotes the Euclidean distance, which computes the sum of the squares of the differences between the label value, $q_i$, and the predicted value, $p_i$, where $N$ indicates the total number of samples.

\section{Experiments}

\subsection{Dataset}
To estimate HR instantaneously, our strategy is to extract the feature image of every second of the video sequence and predict the HR from the feature image. In other words, consecutive video frames of one second of video are input into the feature extraction module, where one corresponding feature image is the output to be used to predict the HR. Therefore, the number of feature images extracted from a whole video is the total number of frames divided by the frame rate. To estimate the HR under practical conditions, a challenging dataset is used in this paper and is introduced as follows.

The MMSE-HR dataset is a subset provided by the MMSE database \cite{zhang2016multimodal}, which was collected for challenging HR estimation. There are 40 participants with diverse ethnic ancestries. 102 RGB videos are recorded, and the length of each video is between 30 seconds to 1 minute. The frame rate is 25 fps, and the resolution of each 2D texture RGB image is 1040$\times$1392 pixels. HR is collected by a contact sensor working at a sample rate of 1 KHz. Since our experiment needs the average HR of every second to be the label of each feature image, the mean of the 1000 HR values within a second is calculated and used in the HR estimation. Each video of the dataset is input into the modules described in Section \ref{face_detect} and Section \ref{feature_extract}, which ultimately produces 5839 feature images for the whole dataset.

\begin{figure}[h!]
    \centering
    \includegraphics[width=0.5\textwidth, right]{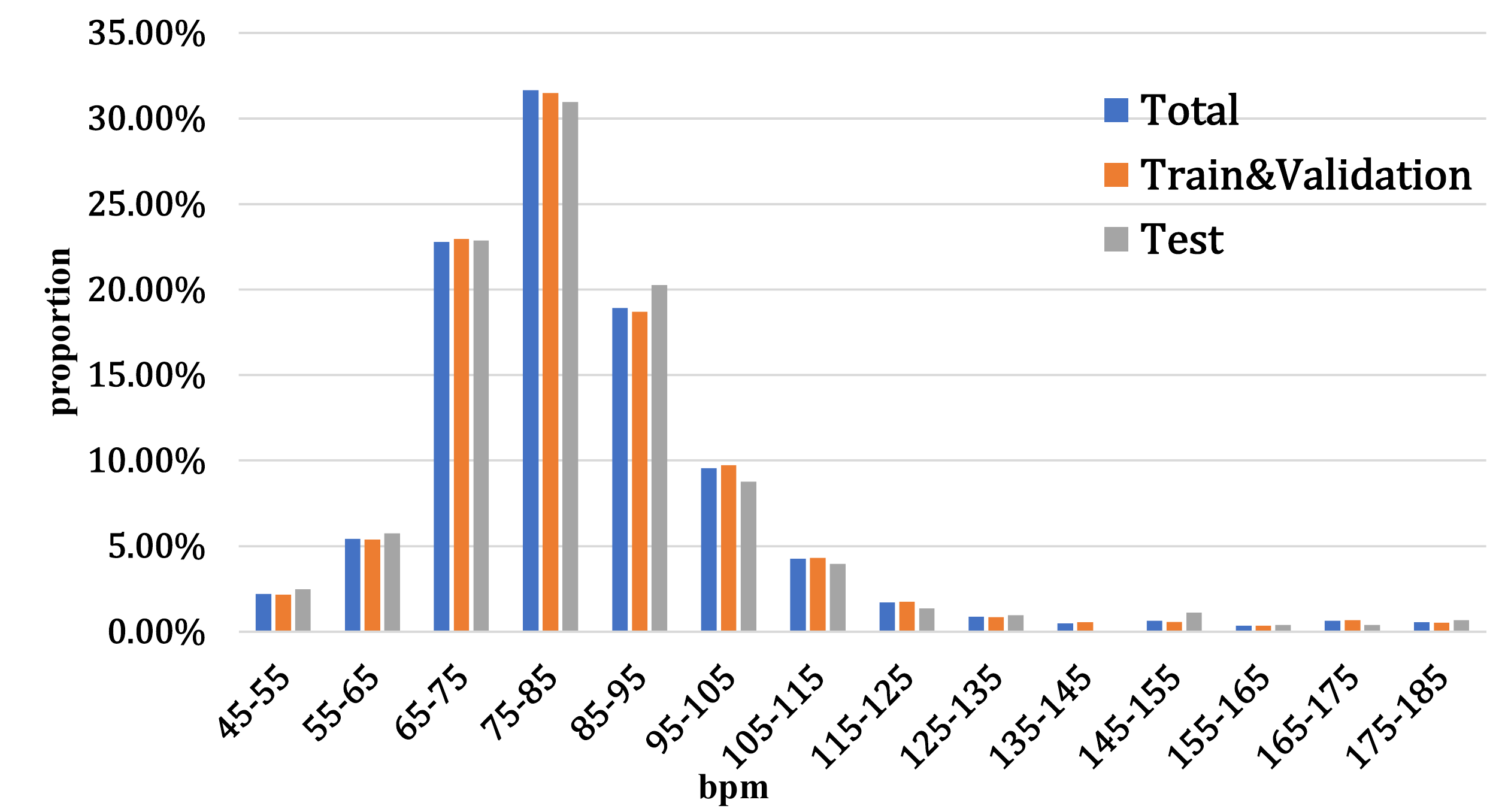}
    \caption{Heart rate proportions from different datasets. The blue bar denotes the HR data from the MMSE-HR dataset \cite{zhang2016multimodal}, the orange bar denotes the HR data from the training and validation dataset, and the gray bar denotes the HR data from the testing dataset.}
    \label{HR_proportion}
\end{figure}

To evaluate the HR data diversity of the dataset, the HR distribution is obtained by calculating the proportion for each range, which is shown in Fig. \ref{HR_proportion}, where the blue bar represents the total HR distribution of the dataset. For the training and testing tasks for HR estimation, 730 feature images are randomly selected as the testing dataset, and the remaining images are used as the training and validation dataset. The HR distributions of these two subsets are also illustrated in Fig. \ref{HR_proportion}, where the orange bar denotes the training and validation dataset and the gray bar denotes the testing dataset. The mean value of the total HR data is 84.28, and the variance is 355.9. The proportion of the total HR data that are between 65 bpm and 95 bpm is 73\%, and the two subsets show similar distributions compared with the full dataset. For the training and validation dataset, the mean of the HR is 84.32, and the variance is 355.6, while for the testing dataset, the mean of the HR is 84.0, and the variance is 357.6.

\subsection{Evaluation Metrics}
In our experiments, five evaluation metrics are used and explained as below. All the evaluation metrics are based on the HR error, which is the difference between the predicted HR and the ground truth.
\begin{equation}
H_{e}(i)=H_{p}(i)-H_{gt}(i)
\end{equation}
where $H_{e}$ denotes the measurement error, $H_{p}$ denotes the predicted HR and $H_{gt}$ denotes the ground-truth HR.
\subsubsection{Mean of $H_{e}$} 
The mean of the measure error is the average value of $H_{e}$:
\begin{equation}
M_{e}=\frac{1}{N}\sum_{i=1}^{N}H_{e}(i)
\end{equation}
where $M_{e}$ denotes the mean of the measure error and $N$ denotes the number of measurements. 
\subsubsection{Standard deviation} is used to quantify the amount of variation in a set of data values. In our experiments, the standard deviation is used to evaluate the dispersion of $H_{e}$:
\begin{equation}
SD_{e}=\sqrt{\frac{\sum_{i=1}^{N}(H_{e}(i)-M_{e})^2}{N}}
\end{equation}
where $SD_{e}$ denotes the standard deviation. $SD_{e}\in[0, \infty)$, where a lower value indicates that the data points of $H_{e}$ tend to be closer to the mean value.
\subsubsection{Root-mean-square error} is used to measure the differences between the values predicted by an estimator and the observed values. It is sensitive to outliers.
\begin{equation}
RMSE=\sqrt{\frac{\sum_{i=1}^{N}(H_{p}(i)-H_{gt}(i))^2}{N}}
\end{equation}
where $RMSE$ denotes the root-mean-square error. $RMSE\in[0, \infty)$ and a lower value indicates that there are fewer outliers in $H_{e}$.
\subsubsection{Mean absolute percentage error} is a measure of the prediction accuracy, which usually expresses the accuracy as a percentage.
\begin{equation}    
M_{eRate}=\frac{1}{N}\sum_{i=1}^{N}\frac{|H_{e}(i)|}{H_{gt}(i)}
\end{equation}
where $M_{eRate}$ denotes the mean absolute percentage error and $M_{eRate}\in[0,+\infty)$, where a lower value indicates that the predicted value is closer to the ground-truth value.
\subsubsection{Pearson's correlation} is used to measure the linear correlation between the predicted HR and the ground-truth value.
\begin{equation}    
\rho=\frac{\sum_{i=1}^{N}(H_{gt}(i)-\overline{H}_{gt})(H_{p}(i)-\overline{H}_{p})}{\sqrt{\sum_{i=1}^{N}(H_{gt}(i)-\overline{H}_{gt})^2}\sqrt{\sum_{i=1}^{N}(H_{p}(i)-\overline{H}_{p})^2}}
\end{equation}
where $\rho$ denotes Pearson's correlation, $\overline{H}_{gt}$ denotes the mean value of the ground truth and $\overline{H}_{p}$ denotes the mean value of the predicted HR. $\rho\in[-1, +1]$, where 1 indicates total positive linear correlation and -1 indicates total negative correlation.

\subsection{Experiment Design}
In this part, three experiments are introduced and used to evaluate our approach.
\subsubsection{Experiment 1}
\begin{figure}[h!]
    \centering
    \includegraphics[width=0.5\textwidth, left]{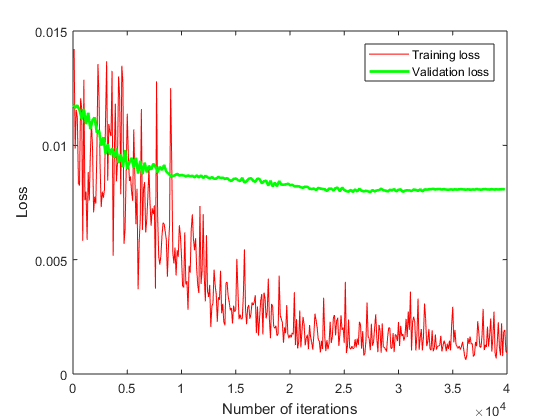}
    \caption{Training behavior of the neural network. The red curve denotes the training loss, the green curve(bold) denotes the validation loss. (The loss curves are not smoothed.)}
    \label{training process}
\end{figure}

The convolutional neural network is used for HR estimation in our method. In this experiment, CNN is evaluated on the testing dataset. As discussed previously, 730 feature images extracted from the whole dataset are randomly chosen to be the testing data, and the rest of the feature images are used in the training and validation steps. The training process is presented in Fig. \ref{training process}. The validation curve is converged after $1.5\times 10^4$ iterations.

\subsubsection{Experiment 2}
In this experiment, the proposed approach is compared with the state-of-the-art methods \cite{tulyakov2016self} \cite{li2014remote} \cite{de2013robust} and evaluated on the challenging MMSE-HR dataset \cite{zhang2016multimodal}. Following \cite{tulyakov2016self}, all the video sequences in this dataset are used for average HR estimation. For each video sequence, in each second, the HR is predicted first by using our approach, and the mean HR is then calculated as our final result.

\subsubsection{Experiment 3}To evaluate the proposed method's ability to estimate HR instantaneously, comparison experiments on short-time HR prediction are conducted. In this experiment, the proposed approach is compared with the same methods as those considered in \textit{Experiment 2}, except for \cite{li2014remote}, which cannot estimate HR instantaneously.  Following \cite{tulyakov2016self}, 20\% of the recorded sequences that have very strong HR variation were selected. Each sequence is split into non-overlapping windows of 4, 6 and 8 seconds. Then, the average HR is calculated for each non-overlapping window. Comparison experiments are conducted independently on each window size. 

In addition, all the experiments are conducted under the Windows 10 platform, and the program was developed by using C++. The convolutional neural network (Section III-C) is run on the deep learning platform of Caffe  [18].  The  level  of  the  Gaussian  pyramid in  the spatial  decomposition  is  set  to 4,  and  the  low  cut-off frequency of the ideal temporal bandpass filter (Section III-B) is set to 0.75 Hz and the high component to 4.0 Hz. The functions applied in the feature extraction module such as: down sampling, FFT, IFFT are implemented by using OpenCV.

\section{Results and Analysis}

\subsection{Visualization of Short-Time HR Estimation}
\begin{figure}[h!]
    \centering
    \includegraphics[width=0.5\textwidth, right]{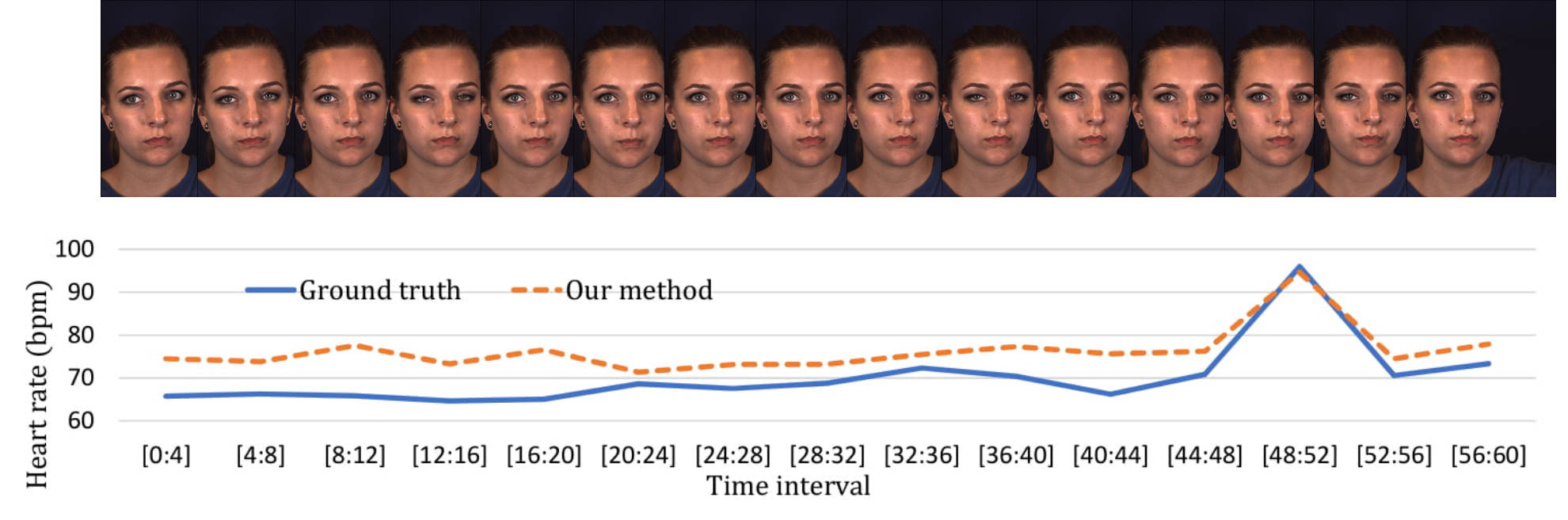}\\
    \includegraphics[width=0.5\textwidth, right]{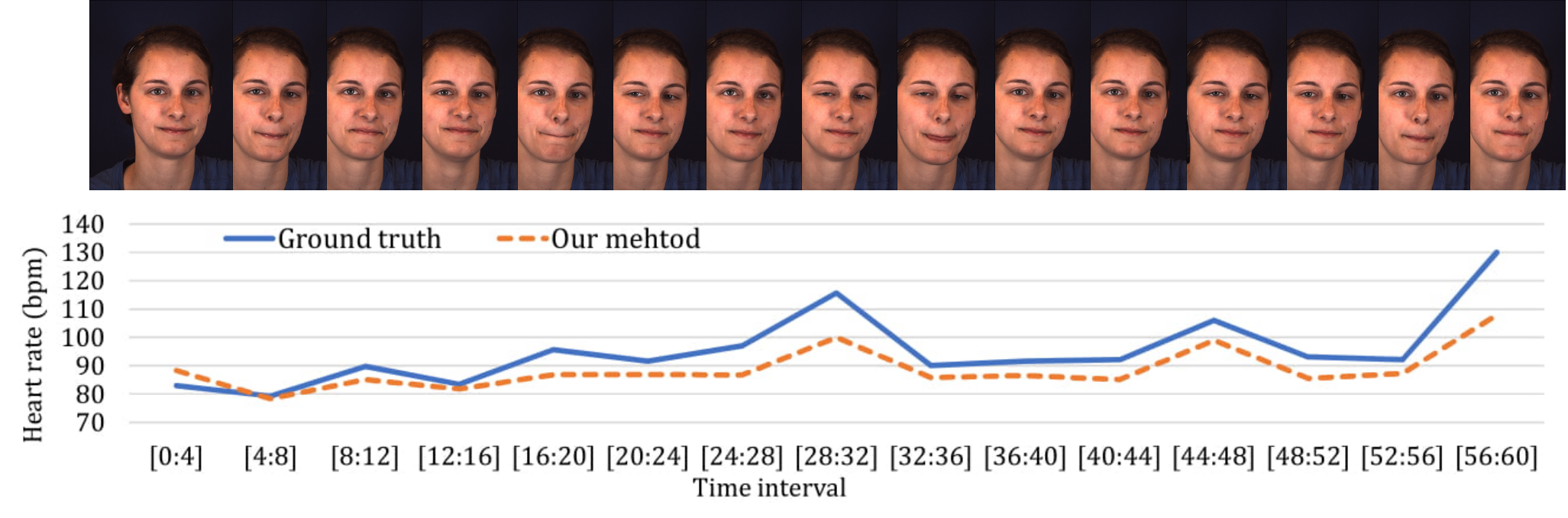}\\
    \includegraphics[width=0.5\textwidth, right]{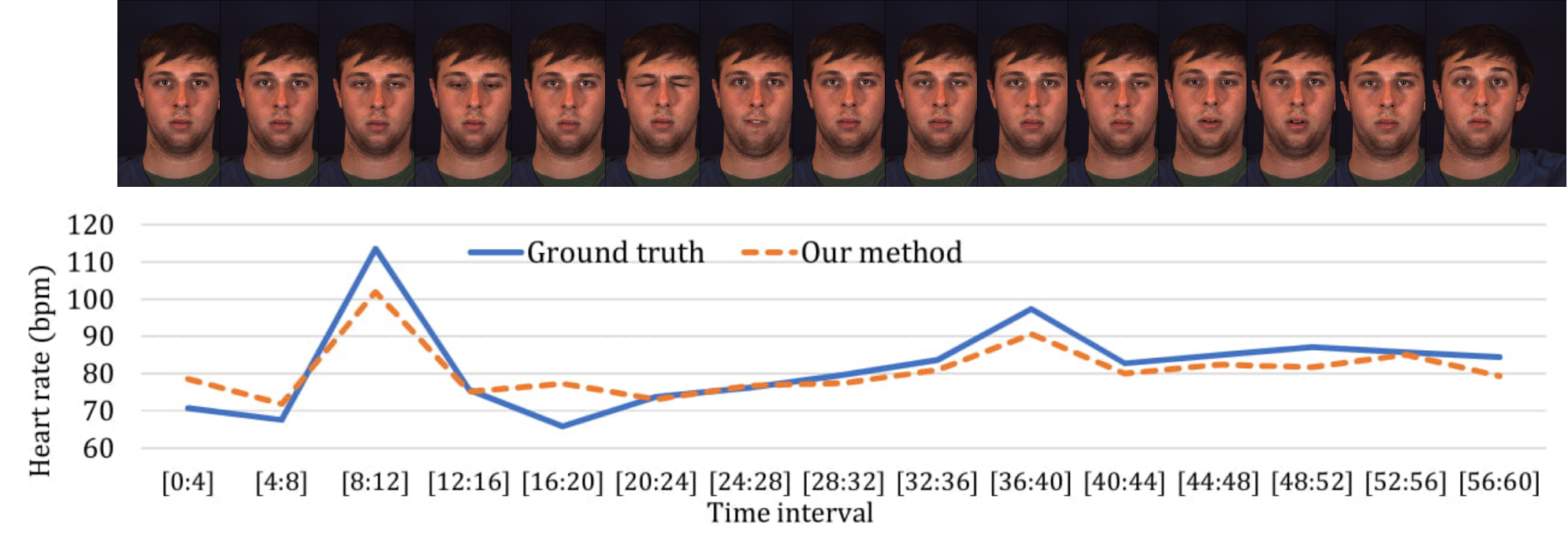}
    \caption{HR estimation of three sequences with a window size of 4 s. The blue line denotes the ground-truth value and the orange line denotes the predicted result by using our method. The frames represent the subject's facial expression; the corresponding HR is shown below.}
    \label{st1}
\end{figure}

To show the performance of the proposed approach for short-time HR estimation, visualizations of the processing of three challenge sequences with a window size of 4s are demonstrated. For each video sequence, the mean of the ground-truth values within 4s is calculated as the final ground-truth value, and one frame is selected from every 4s video sequence as the representative of the subject's facial expression during that time interval, as shown in Fig. \ref{st1}, where the horizontal axis represents the time interval and the vertical axis represents the HR value.

From the first result in Fig. \ref{st1}, the ground truth shows low-frequency HR at the beginning, and a sudden increase and decrease are shown at the end. For the monotonicity of the ground truth, there are 8 increasing stages and 6 decreasing stages, of which 5 are falsely predicted, but all 5 changes are small with a difference of less than 5 bpm. In the second result, the ground truth shows large waves throughout the time interval, while 4 changes are falsely predicted, one of which is larger than 5 bpm. In the third result, 4 changes are falsely predicted, two of which are larger than 5 bpm. Overall, as shown in the figure, the predicted results always have a similar trend to the ground truth; 69\% of the HR changes are correctly predicted, and among the falsely predicted samples, 76.9\% of them have a difference of less than 5 bpm. For these small changes, the predicted results may be influenced by the cheek muscle movements, which leads to a result with large changes.

\subsection{Evaluation of the CNN HR Estimator}

\begin{figure}[h!]
    \centering
    \includegraphics[width=0.5\textwidth, right]{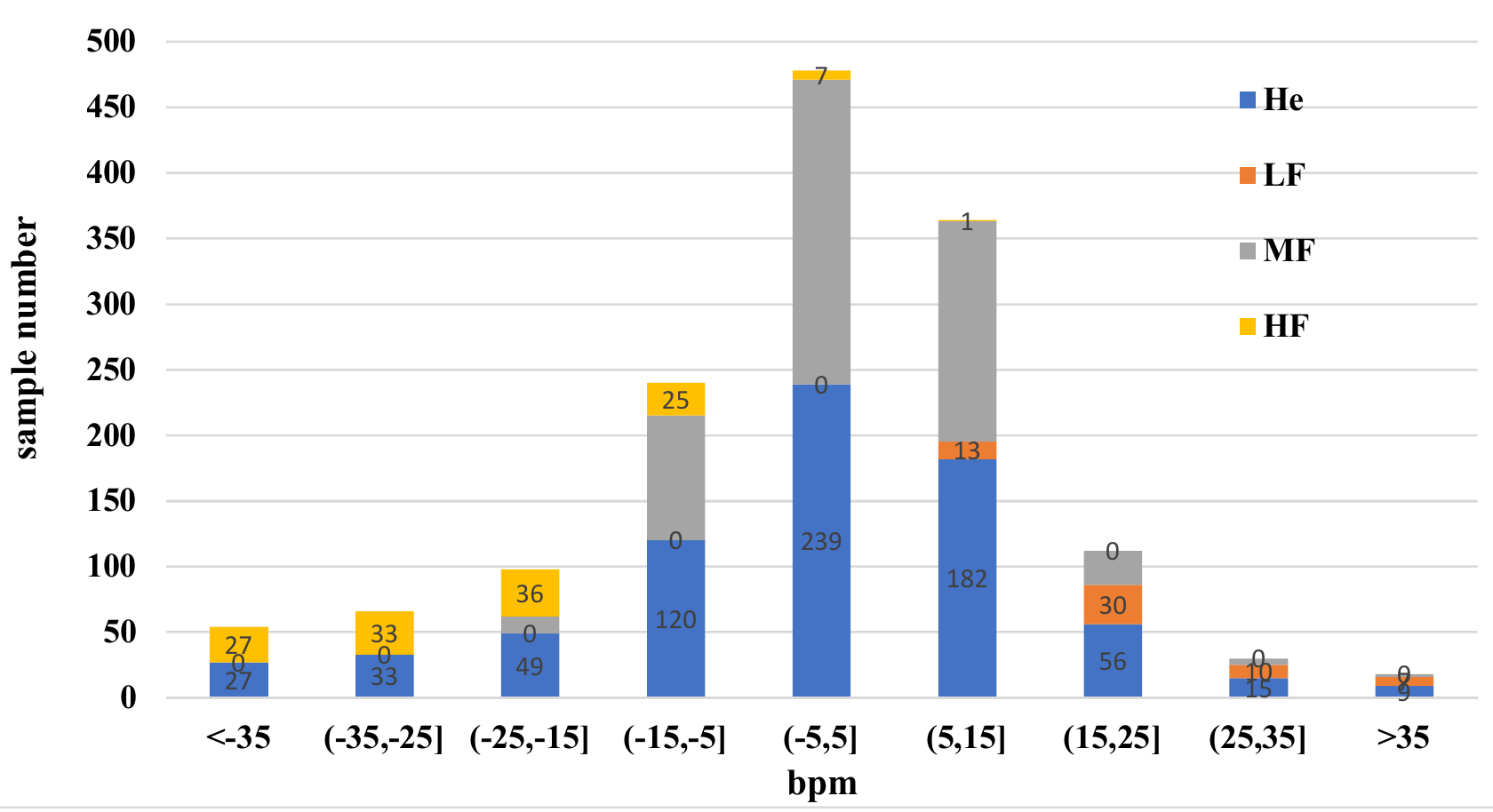}
    \caption{HR error distribution on the testing dataset. The horizontal axis denotes the range of $H_{e}$, and the vertical axis denotes the number of samples. The blue bar represents $H_{e}$, the orange bar represents the LF part (45-65 bpm), the gray bar represents the MF part (65-95 bpm) and the yellow bar represents the HF part (95-185 bpm).}
    \label{He_proportion}
\end{figure}

This result is used to evaluate the CNN heart rate estimator that is obtained in Experiment 1. To demonstrate the differences between the labels and the predicted values, which are defined as Eq.(3), the error is calculated for each sample. Additionally, to evaluate the accuracy of the estimator on various frequency bands, the testing data are divided into three bands according to the HR distribution in the full dataset (Fig. \ref{HR_proportion}). The ranges $[45, 65)$, $[65, 95)$, and $[95, 185]$ are defined as the low-frequency (LF) part, medium-frequency (MF) part and high-frequency (HF) part, respectively. The numbers of samples in each part are 60, 541 and 129, respectively. The result of this experiment is shown in Fig. \ref{He_proportion}, where the LF, MF, and HF parts are represented by orange, gray and yellow bars, respectively, and $H_{e}$ is denoted by a blue bar.

As shown in Fig. \ref{He_proportion}, the error between the labels and the predicted values are distributed in 9 sections, which is demonstrated on the horizontal axis, while the vertical axis represents the proportion. The mean of $H_{e}$ is -1.25, and the variance is 311.25. There are 541 samples with an absolute $H_{e}$ within 15 bpm, which indicates that 74.13\% of the test data are well estimated. In this range, as shown in Fig. \ref{He_proportion}, most of the samples are from the MF part, which takes up 91.5\%, while the LF and HF parts take up 2.4\% and 6.1\%, respectively. In contrast, for an absolute $H_{e}$ of more than 15, most of the samples are from either the HF or LF part. Therefore, for $H_{e}<-15$, the HF part takes up 88\%, while MF part takes up 12\%. For $H_{e}\geq15$, the LF part takes up 58.8\%, while the MF part takes up 41.2\%. This phenomenon is due to the scarcity of the LF and HF samples in the training data, which leads to lower accuracy for these two parts. For the LF part, 78.3\% are predicted to be much higher than the ground truth with $He>15$. For the HF part, 74.4\% are predicted to be much lower than the ground truth with $H_{e}<-15$. However, only 8.5\% of the MF part is predicted with lower accuracy, according to $|H_{e}|>15$. Comparing the percentages of the parts that are predicted with higher HR error indicates that if a frequency part takes up a larger proportion of the dataset, then the predicted result of this part has higher accuracy than the others, which also indicates that the diversity of the HR dataset is important. 

\subsection{Average HR Prediction on the MMSE-HR Dataset}

This result is obtained from Experiment 2, which is used to compare the performances in terms of average HR prediction with other methods on the same dataset. The result is evaluated by using five metrics, which were defined earlier and commonly used in existing research. The result is shown in Table \ref{table:1}. The performance comparison results between \cite{li2014remote,de2013robust,tulyakov2016self} are taken from \cite{tulyakov2016self}.

\begin{table}[h!]
\centering
\caption{Average HR Prediction: comparison on MMSE-HR dataset}
\resizebox{0.48\textwidth}{!}{
\begin{tabular}{c|c|c|c|c}
\hline
 Method & $M_{e}(SD_{e})$ & $RMSE$ & $M_{eRate}$ & $\rho$\\
\hline
Li \textit{et al.} \cite{li2014remote} & 11.56(20.02) & 19.95 & 14.64\% & 0.38\\
\hline
Haan \textit{et al.} \cite{de2013robust} & 9.41(14.08) & 13.97 & 12.22\% & 0.55\\
\hline
Tulyakov \textit{et al.} \cite{tulyakov2016self} & 7.61(12.24) & 11.37 & 10.84\% & 0.71\\
\hline
Ours & -1.17(\textbf{6.85}) & \textbf{6.95} & \textbf{6.55}\% & \textbf{0.98}\\
\hline
\end{tabular}}
\label{table:1}
\end{table}

As shown in Table \ref{table:1}, compared with existing methods, $SD_{e}$ of our result is reduced to 6.85, which shows that the HR error tends to be closer to the mean value. $RMSE$ is reduced to 6.95, which indicates that the magnitudes of the errors in the predictions are decreased. The results for $M_{eRate}$ demonstrate that the prediction accuracy is improved to 6.55\% when using our approach. In addition, $\rho$ is improved to 0.98, which is closer to 1; thus, the linear correlation between ground truth and predicted HR values is stronger. Overall, in all aspects, our approach outperforms existing methods \cite{tulyakov2016self} \cite{li2014remote} \cite{de2013robust} on average HR prediction. Comparing each metric value among all the methods, it is shown that as $\rho$ increases, all the other values decrease.

\subsection{Short-Time HR Estimation}

This result is obtained from Experiment 3, which is used to evaluate the performance on short-time HR estimation and compare it with those of other methods. Five metrics are also used here to evaluate the results, which are shown in Table \ref{table:2}. The performance measures for \cite{de2013robust} is taken from \cite{tulyakov2016self}.

\begin{table}[h!]
\centering
\caption{Short-Time HR Prediction Performance: comparison on MMSE-HR dataset}
\resizebox{0.48\textwidth}{!}{
\begin{subtable}{0.4\textwidth}
\label{table:a}
\begin{tabular}{c|c|c|c|c}
\hline
Method & $M_{e}(SD_{e})$ & $RMSE$ & $M_{eRate}$ & $\rho$\\
\hline
Haan \textit{et al.} \cite{de2013robust} & -1.85(15.77) & 15.83 & 9.92\% & 0.67\\
\hline
Tulyakov \textit{et al.} \cite{tulyakov2016self} & 2.12(11.51) & 11.66 & 9.15\% & 0.78\\
\hline
Ours & -1.31(\textbf{8.19}) & \textbf{8.30} & \textbf{6.93}\% & \textbf{0.93}\\
\hline
\end{tabular}
\caption{With Window Size 4s}
\end{subtable}}

\resizebox{0.48\textwidth}{!}{
\begin{subtable}{0.4\textwidth}
\label{table:b}
\begin{tabular}{c|c|c|c|c}
\hline
Method & $M_{e}(SD_{e})$ & $RMSE$ & $M_{eRate}$ & $\rho$\\
\hline
Haan \textit{et al.} \cite{de2013robust} & -2.21(19.21) & 19.27 & 11.81\% & 0.33\\
\hline
Tulyakov \textit{et al.} \cite{tulyakov2016self} & 0.32(8.29) & 8.27 & 7.30\% & 0.80\\
\hline
Ours & -1.29(\textbf{7.53}) & \textbf{7.64} & \textbf{6.74}\% & \textbf{0.95} \\
\hline
\end{tabular}
\caption{With Window Size 6s}
\end{subtable}}

\resizebox{0.48\textwidth}{!}{
\begin{subtable}{0.4\textwidth}
\label{table:c}
\begin{tabular}{c|c|c|c|c}
\hline
Method & $M_{e}(SD_{e})$ & $RMSE$ & $M_{eRate}$ & $\rho$\\
\hline
Haan \textit{et al.} \cite{de2013robust} & 0.81(11.49) & 11.46 & 8.60\% & 0.63\\
\hline
Tulyakov \textit{et al.} \cite{tulyakov2016self} & 1.62(9.67) & 9.76 & 7.52\% & 0.71\\
\hline
Ours & -1.24(\textbf{7.24}) & \textbf{7.34} & \textbf{6.58}\% & \textbf{0.96}\\
\hline
\end{tabular}
\caption{With Window Size 8s}
\end{subtable}}
\label{table:2}
\end{table}

As shown in Table \ref{table:2}(a) (b) (c), compared with the other methods, for each window size, $SD_{e}$, $RMSE$ and $M_{eRate}$ from our approach are all less than those from existing methods, which indicates that the variations of the predicted HR distribute over a narrower range with a higher accuracy. $\rho$ is also improved by our approach for each window size, which shows a stronger linear correlation between the predicted values and the ground truth. Haan's \cite{de2013robust} method has the worst performance on the experiment with a window size of 6 s, with $\rho$ equal to 0.33. Conversely, Tulyakov's \cite{tulyakov2016self} method performs best for a window size of 6 s. However, from the results of our methods in experiments with different window sizes, as the window size grows, $SD_{e}$, $RMSE$ and $M_{eRate}$ decrease while $\rho$ increases, which shows that the performance improves steadily in all aspects. Since our approach is to estimate HR for every second and calculate the mean value over a pre-defined time interval, as the time interval grows, the accuracy increases.

\subsection{Run Time}
To calculate the run time, the full framework is divided into two parts. The first part includes video capture, frame extraction, face detection, tracking, ROI cropping and feature image extraction. The second part is the HR estimator. Part 1 runs at 110 fps, and part 2 runs at 290 fps. For the training phase, there are 20 images in a batch, one iteration takes around 0.6s. The running times are measured using a conventional laptop with an Intel Core i5-7300HQ CPU and 8.0 GB RAM. Thus, the proposed approach runs fast enough to be used for real-time HR estimation.

\subsection{Further Discussion}
As introduced before, the proposed method is based on the ROI extraction, however, the face detection has limitations in some situations. In realistic conditions, subject's head can rotate randomly in front of the camera, while in the proposed approach, only faces that rotate within a certain angle range can be detected. Specifically, the subject's head can rotate around yaw from -90 degree to 90 degree, around roll from -20 degree to 20 degree and around pitch from -20 degree to 20 degree. For the positions that 68 landmarks are not identified, there is no ROI extracted and no corresponding HR estimated. The whole process only works when consecutive faces are detected within 1 s.

We also evaluate the proposed method on the MAHNOB-HCI dataset \cite{soleymani2012multimodal}. There are 27 participants (12 males and 15 females) involved, 527 videos are recorded and can be used totally. In our experiments, same as \cite{tulyakov2016self}, 30 seconds interval (frame 306 to 2135) is extracted from each video sequence. The second channel (EXG2) of ECG signal is used to obtain the ground truth. Half of the video sequences are randomly chosen to extract the feature images and trained. A comparison result is shown in Table. \ref{table:hci}, where all the video sequences are tested and the comparison results for \cite{poh2010non,poh2011advancements,balakrishnan2013detecting,li2014remote,de2013robust,tulyakov2016self} are taken from \cite{tulyakov2016self}. Apparently, the proposed approach also shows better performance on the MAHNOB-HCI dataset. 

\begin{table}[h!]
\centering
\caption{Average HR Prediction: comparison results on MAHNOB-HCI dataset}
\resizebox{0.48\textwidth}{!}{
\begin{tabular}{c|c|c|c|c}
\hline
 Method & $M_{e}(SD_{e})$ & $RMSE$ & $M_{eRate}$ & $\rho$\\
\hline
Poh \textit{et al.} \cite{poh2010non} & -8.95(24.3) & 25.9 & 25.0\% & 0.08\\
\hline
Poh \textit{et al.} \cite{poh2011advancements} & 2.04(13.5) & 13.6 & 13.2\% & 0.36\\
\hline
Balakrishnan \textit{et al.} \cite{balakrishnan2013detecting} & -14.4(15.2) & 21.0 & 20.7\% & 0.11\\
\hline
Li \textit{et al.} \cite{li2014remote} & -3.30(6.88) & 7.62 & 6.87\% & 0.81\\
\hline
Haan \textit{et al.} \cite{de2013robust} & 4.62(6.50) & 6.52 & 6.39\% & 0.82\\
\hline
Tulyakov \textit{et al.} \cite{tulyakov2016self} & 3.19(5.81) & 6.23 & 5.93\% & 0.83\\
\hline
Ours & -1.68(\textbf{2.79}) & \textbf{3.26} & \textbf{3.67}\% & \textbf{0.95}\\
\hline
\end{tabular}}
\label{table:hci}
\end{table}

\section{Conclusion}

In this paper, a new framework is introduced for contactless HR estimation from facial videos under realistic conditions. Different from the traditional HR estimation approach, which usually extracts a signal related to HR and uses power spectrum density analysis to estimate the average HR, a convolutional neural network is used to estimate HR in the proposed approach. Instead of applying a series of filters to clean the underlying signal, HR is directly estimated from a feature image that is obtained by using spatial decomposition and temporal filtering, which decreases both the computational complexity and the processing time.

The results of testing the trained model on the testing dataset show that 74.13\% of the data in the testing dataset are well estimated, while the rest have large error, which is due to the lack of high- and low-frequency components in the dataset. In addition, comparison experiments are conducted on the MMSE-HR dataset for both average HR estimation and short-time HR estimation, and the results show that the proposed approach achieves higher accuracy than other methods. 

Furthermore, our approach can be improved in the future. For instance, the HR diversity of the dataset can be improved. As shown in Fig. \ref{HR_proportion}, the low-frequency and high-frequency parts take up no more than 30\% of the data, which affects the estimation performance on these two parts. Therefore, a larger dataset with well-proportioned HR data can be collected in future work. 

\bibliographystyle{unsrt}
\bibliography{references}

\begin{IEEEbiography}[{\includegraphics[width=1in,height=1.25in,clip,keepaspectratio]{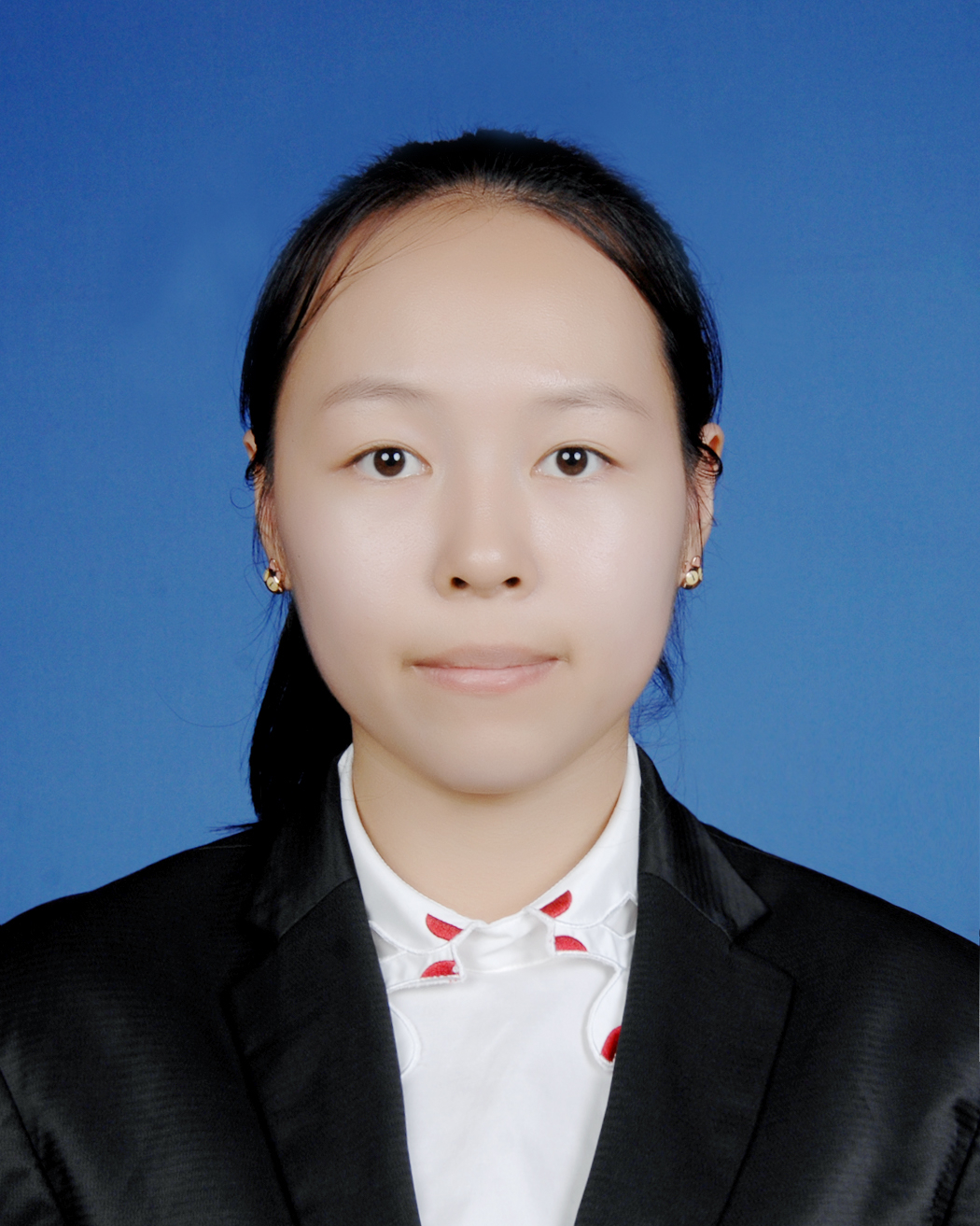}}]{Ying Qiu}
received the B.Eng. degree in measurement and control engineering from Southwest Jiao Tong University (China), in 2016. She is currently pursuing her M.Sc. degree with the Multimedia Computing Research Laboratory at the University of Ottawa. Her research interests include machine learning and computer vision.
\end{IEEEbiography}

\begin{IEEEbiography}[{\includegraphics[width=1in,height=1.25in,clip,keepaspectratio]{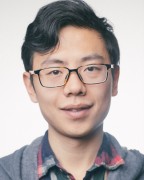}}]{Yang Liu}
received M.Sc. degree with the Multimedia Computing Research Laboratory from University of Ottawa and the B.Eng. degree in information engineering from Beijing Technology and Business University (China), in 2017 and 2014. He is currently working in SPORTLOGIQ as Computer Vision Researcher. His research interests include computer vision and computer graphic.

\end{IEEEbiography}

\begin{IEEEbiography}[{\includegraphics[width=1in,height=1.25in,clip,keepaspectratio]{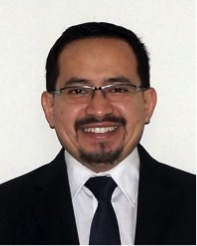}}]{Juan Arteaga-Falconi} is currently a Ph.D Candidate in Electrical and Computer Engineering at the University of Ottawa, Ottawa, ON, Canada. He received the Engineering degree in Electronics from the Politecnica Salesiana University, Cuenca, AZ, Ecuador, in 2008, and the M.A.Sc. Degree in Electrical and Computer Engineering from the University of Ottawa in 2013.
From 2008 to 2011 he was with SODETEL Co. Ltd., Cuenca, AZ, Ecuador, where he was a Co-Founder and served as General Manager. He  joined the MCRLab at the University of Ottawa in 2012 and currently he is a Teaching Assistant at the same university. His research interests are: Biometrics, Signal Processing, System Security, Machine Learning and Internet of Things.
Mr. Arteaga-Falconi has served as Treasurer in the IEEE ExCom of the Ecuadorian section from 2010 to 2012. He has received the 2011 and 2013 SENESCYT Ecuadorian Scholarship for graduate studies.

\end{IEEEbiography}

\begin{IEEEbiography}[{\includegraphics[width=1in,height=1.25in,clip,keepaspectratio]{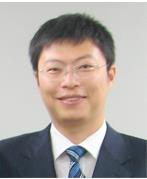}}]{Haiwei Dong}
(M'12-SM'16)  received the Dr.Eng. degree in computer science and systems engineering from Kobe University, Japan and M.Eng. degree in control theory and control engineering from Shanghai Jiao Tong University, P.R.China in 2008 and 2010, respectively. He is currently a Research Scientist in the University of Ottawa. Prior to that, he was a Post-Doctoral Fellow at New York University, a Research Associate at the University of Toronto, a Research Fellow (PD) at the Japan Society for the Promotion of Science, a Science Technology Researcher at Kobe University, and a Science Promotion Researcher at the Kobe Biotechnology Research and Human Resource Development Center. His research interests include robotics, multimedia, and deep learning.
\end{IEEEbiography}

\begin{IEEEbiography}[{\includegraphics[width=1in,height=1.25in,clip,keepaspectratio]{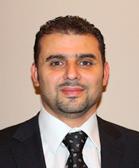}}]{Abdulmotaleb El Saddik}
(M'01-SM'04-F'09) is currently a Distinguished University Professor and the University Research Chair with the School of Electrical Engineering and Computer Science, University of Ottawa. He has authored and co-authored four books and more than 550 publications and chaired more than 40 conferences and workshops. His research focus is on multimodal interactions with sensory information in smart cities. He has received research grants and contracts totaling more than \$18 M. He has supervised over 120 researchers and received several international awards, among others, are the ACM Distinguished Scientist, Fellow of the Engineering Institute of Canada, Fellow of the Canadian Academy of Engineers, and also received the IEEE I\&M Technical Achievement Award and the IEEE Canada Computer Medal.
\end{IEEEbiography}
\end{document}